\documentclass[
  manuscript=commentary,
]
{cup-modified}
\usepackage[utf8]{inputenc}
\usepackage{setspace}
\usepackage{natbib}
\setcitestyle{authoryear,open={(},close={)}}

\title{Beyond the limitations of any imaginable mechanism: large language models and psycholinguistics}
\author{Conor Houghton%
}
\affiliation{Department of Computer Science, University of Bristol, UK%
}
\email[CH]{conor.houghton@bristol.ac.uk
}
\author{Nina Kazanina
}
\affiliation{School of Psychological Sciences, University of Bristol, UK%
}
\alsoaffiliation{International Laboratory of Social Neurobiology, Institute for Cognitive Neuroscience, National Research University Higher School of Economics, HSE University, Moscow, Russia.
}
\author{Priyanka Sukumaran%
}
\affiliation{School of Psychological Sciences, University of Bristol, UK%
}

\begin{document}

\begin{abstract}
 Large language models are not detailed models of human linguistic
 processing. They are, however, extremely successful at their primary
 task: providing a model for language. For this reason and because
 there are no animal models for language, large language models are
 important in psycholinguistics: they are useful as a practical tool,
 as an illustrative comparative, and philosophically, as a basis for
 recasting the relationship between language and
 thought.\\ \\ \textbf{This is a commentary on \cite{BowersEtAl2023}}.
\end{abstract}

Neural-network models of language are optimized to solve practical
problems such as machine translation. Currently, when these large
language models (LLMs) are interpreted as models of human linguistic
processing they have similar shortcomings to those that deep neural
network have as models of human vision. Two examples can illustrate
this. First, LLMs do not faithfully replicate human behaviour on
language tasks
\citep{MarvinLinzen2018,KuncoroEtAl2018,LinzenLeonard2018,MitchellEtAl2019}. For
example, an LLM trained on a word-prediction task shows similar error
rates to humans overall on long-range subject-verb number agreement
but errs in different circumstances: unlike humans, it makes more
mistakes when sentences have relative clauses
\citep{LinzenLeonard2018}, indicating differences in how grammatical
structure is represented. Second, the LLMs with better performance on
language tasks do not necessarily have more in common with human
linguistic processing or more obvious similarities to the brain. For
example, Transformers learn efficiently on vast corpora and avoid
human-like memory constraints but are currently more successful as
language models than recurrent neural networks such as the
Long-Short-Term-Memory LLMs \citep{DevlinEtAl2018,BrownEtAl2020},
which employ sequential processing, as humans do, and can be more
easily compared to the brain.

Furthermore, the target article suggests that, more broadly, the brain
and neural networks are unlikely to resemble each other because
evolution differs in trajectory and outcome from the optimization used
to train a neural network. Generally, there is an unanswered question
about which aspects of learning in LLMs are to be compared to the
evolution of our linguistic ability and which to language learning in
infants but in either case, the comparison seems weak. LMMs are
typically trained using a next-word prediction task; it is unlikely
our linguistic ability evolved to optimize this and next-word
prediction can only partly describe language learning: for example,
infants generalize word meanings based on shape
\citep{LandauSmithJones1988} while LLMs lack any broad conceptual
encounter with the world language describes.

In fact, it would be peculiar to suggest that LLMs are models of the
neural dynamics that support linguistic processing in humans; we
simply know too little about those dynamics. The challenge presented
by language is different to that presented by vision: language lacks
animal models and debate in psycholinguistics is occupied with broad
issues of mechanisms and principles, whereas visual neuroscience often
has more detailed concerns. We believe that LLMs have a valuable role
in psycholinguistics and this does not depend on any precise mapping
from machine to human. Here we describe three uses of LLMs:
(\textbf{1}) the \textbf{practical}, as a tool in experimentation;
(\textbf{2}) the \textbf{comparative}, as an alternate example of
linguistic processing and (\textbf{3}) the \textbf{philosophical},
recasting the relationship between language and thought.

(\textbf{1}): An LLM models language and this is often of
\textbf{practical} quantitative utility in experiment. One
straight-forward example is the evaluation of \textsl{surprisal}: how
well a word is predicted by what has preceded it. It has been
established that reaction times,
\citep{FischlerBloom1979,Kleiman1980}, gaze duration,
\citep{RaynerWell1996}, and EEG responses,
\citep{DambacherEtAl2006,FrankEtAl2015}, are modulated by surprisal,
giving an insight into prediction in neural processing. In the past,
surprisal was evaluated using $n$-grams, but $n$-grams become
impossible to estimate as $n$ grows and as such they cannot quantify
long-range dependencies. LLMs are typically trained on a task akin to
quantifying surprisal and are superior to $n$-grams in estimating word
probabilities. Differences between LLM-derived estimates and neural
perception of surprisal may quantify which linguistic structures,
perhaps poorly represented in the statistical evidence, the brain
privileges during processing.

(\textbf{2}): LLMs are also useful as a point of
\textbf{comparison}. LLMs combine different computational strategies,
mixing representations of word properties with a computational engine
based on memory or attention. Despite the clear differences between
LLMs and the brain, it is instructive to compare the performance of
different LLMs on language tasks to our own language ability. For
example, although LLMs are capable of long range number and gender
agreement,
\citep{LinzenDupouxGoldberg2016,GulordavaEtAl2018,BernardyLappin2017,SukumaranEtAl2022},
they are not successful in implementing another long-range rule:
Principle C, \citep{MitchellEtAl2019}, a near-universal property of
languages which depends in its most straight-forward description on
hierarchical parsing. Thus, LLMs allow us to recognize those aspects
of language which require special consideration while revealing others
to be within easy reach of statistical learning.

(\textbf{3}): In the past, \textbf{philosophical} significance was
granted to language as evidence of thought or
personhood. \cite{Turing1950}, for example, proposes conversation as a
proxy for thought and \cite{Chomsky2009} describes Descartes as
attributing the possession of mind to other humans because the human
capacity for innovation and for the creative use of language, is
`beyond the limitations of any imaginable mechanism'. It is
significant that machines are now capable of imitating the use of
language. While machine-generated text still has attributes of
awkwardness and repetition that make it recognizable on careful
reading, it would seem foolhardy to predict these final quirks are
unresolvable or are characteristic of the division between human and
machine. Nonetheless, most of us appear to feel intuitively that LLMs
enact an imitation rather than a recreation of our linguistic ability:
LLMs seem empty things whose pantomime of language is not underpinned
by thought, understanding or creativity. Indeed, even if an LLM were
capable of imitating us perfectly, we would still distinguish between
a loved one and their simulation.

This is a challenge to our understanding of the relationship between
language and thought: either we must claim that, despite recent
progress, machine-generated language will remain unlike human language
in vital respects, or we must defy our intuition and consider machines
to be as capable of thought as we are, or we must codify our intuition
to specify why a machine able to produce language should, nonetheless,
be considered lacking in thought.

\begingroup
\setstretch{0.75}

\endgroup


\begin{thebibliography}{19}
\providecommand{\natexlab}[1]{#1}
\providecommand{\url}[1]{\texttt{#1}} \expandafter\ifx\csname
urlstyle\endcsname\relax \providecommand{\doi}[1]{doi: #1}\else
\providecommand{\doi}{doi: \begingroup \urlstyle{rm}\Url}\fi

\bibitem[Bernardy and Lappin(2017)]{BernardyLappin2017}
J.-P. Bernardy and S.~Lappin.
\newblock Using deep neural networks to learn syntactic agreement.
\newblock \emph{Linguistic Issues in Language Technology}, 2017.

\bibitem[Bowers et~al.(2022)Bowers, Malhotra, Dujmović, Montero, Tsvetkov,
  Biscione, Puebla, Adolfi, Hummel, Heaton, and et~al.]{BowersEtAl2023}
J.~S. Bowers, G.~Malhotra, M.~Dujmović, M.~L. Montero, C.~Tsvetkov,
  V.~Biscione, G.~Puebla, F.~Adolfi, J.~E. Hummel, R.~F. Heaton, and et~al.
\newblock Deep problems with neural network models of human vision.
\newblock \emph{Behavioral and Brain Sciences}, page 1–74, 2022.
\newblock \url{doi.org/10.1017/S0140525X22002813}.

\bibitem[Brown et~al.(2020)Brown, Mann, Ryder, Subbiah, Kaplan, Dhariwal,
  Neelakantan, Shyam, Sastry, Askell, et~al.]{BrownEtAl2020}
T.~Brown, B.~Mann, N.~Ryder, M.~Subbiah, J.~D. Kaplan, P.~Dhariwal,
  A.~Neelakantan, P.~Shyam, G.~Sastry, A.~Askell, et~al.
\newblock Language models are few-shot learners.
\newblock \emph{Advances in Neural Information Processing Systems},
  33:\penalty0 1877--1901, 2020.

\bibitem[Chomsky(1966)]{Chomsky2009}
N.~Chomsky.
\newblock \emph{Cartesian linguistics: A chapter in the history of rationalist
  thought}.
\newblock Cambridge University Press, 1966.

\bibitem[Dambacher et~al.(2006)Dambacher, Kliegl, Hofmann, and
  Jacobs]{DambacherEtAl2006}
M.~Dambacher, R.~Kliegl, M.~Hofmann, and A.~M. Jacobs.
\newblock Frequency and predictability effects on event-related potentials
  during reading.
\newblock \emph{Brain Research}, 1084\penalty0 (1):\penalty0 89--103, 2006.

\bibitem[Devlin et~al.(2018)Devlin, Chang, Lee, and Toutanova]{DevlinEtAl2018}
J.~Devlin, M.-W. Chang, K.~Lee, and K.~Toutanova.
\newblock {BERT}: {Pre-training} of deep bidirectional transformers for
  language understanding.
\newblock \emph{arXiv:1810.04805}, 2018.

\bibitem[Fischler and Bloom(1979)]{FischlerBloom1979}
I.~Fischler and P.~A. Bloom.
\newblock Automatic and attentional processes in the effects of sentence
  contexts on word recognition.
\newblock \emph{Journal of Verbal Learning and Verbal Behavior}, 18\penalty0
  (1):\penalty0 1--20, 1979.

\bibitem[Frank et~al.(2015)Frank, Otten, Galli, and Vigliocco]{FrankEtAl2015}
S.~L. Frank, L.~J. Otten, G.~Galli, and G.~Vigliocco.
\newblock The {ERP} response to the amount of information conveyed by words in
  sentences.
\newblock \emph{Brain and Language}, 140:\penalty0 1--11, 2015.

\bibitem[Gulordava et~al.(2018)Gulordava, Bojanowski, Grave, Linzen, and
  Baroni]{GulordavaEtAl2018}
K.~Gulordava, P.~Bojanowski, E.~Grave, T.~Linzen, and M.~Baroni.
\newblock Colorless green recurrent networks dream hierarchically.
\newblock \emph{arXiv:1803.11138}, 2018.

\bibitem[Kleiman(1980)]{Kleiman1980}
G.~M. Kleiman.
\newblock Sentence frame contexts and lexical decisions: Sentence-acceptability
  and word-relatedness effects.
\newblock \emph{Memory \& Cognition}, 8\penalty0 (4):\penalty0 336--344, 1980.

\bibitem[Kuncoro et~al.(2018)Kuncoro, Dyer, Hale, and Blunsom]{KuncoroEtAl2018}
A.~Kuncoro, C.~Dyer, J.~Hale, and P.~Blunsom.
\newblock The perils of natural behaviour tests for unnatural models: the case
  of number agreement.
\newblock \emph{Poster presented at Learning Language in Humans and in
  Machines, Paris, Fr., July}, 5\penalty0 (6), 2018.

\bibitem[Landau et~al.(1988)Landau, Smith, and Jones]{LandauSmithJones1988}
B.~Landau, L.~B. Smith, and S.~S. Jones.
\newblock The importance of shape in early lexical learning.
\newblock \emph{Cognitive Development}, 3\penalty0 (3):\penalty0 299--321,
  1988.

\bibitem[Linzen and Leonard(2018)]{LinzenLeonard2018}
T.~Linzen and B.~Leonard.
\newblock Distinct patterns of syntactic agreement errors in recurrent networks
  and humans.
\newblock \emph{arXiv:1807.06882}, 2018.

\bibitem[Linzen et~al.(2016)Linzen, Dupoux, and
  Goldberg]{LinzenDupouxGoldberg2016}
T.~Linzen, E.~Dupoux, and Y.~Goldberg.
\newblock Assessing the ability of {LSTMs} to learn syntax-sensitive
  dependencies.
\newblock \emph{Transactions of the Association for Computational Linguistics},
  4:\penalty0 521--535, 2016.

\bibitem[Marvin and Linzen(2018)]{MarvinLinzen2018}
R.~Marvin and T.~Linzen.
\newblock Targeted syntactic evaluation of language models.
\newblock \emph{arXiv:1808.09031}, 2018.

\bibitem[Mitchell et~al.(2019)Mitchell, Kazanina, Houghton, and
  Bowers]{MitchellEtAl2019}
J.~Mitchell, N.~Kazanina, C.~Houghton, and J.~Bowers.
\newblock Do {LSTMs} know about {Principle C?}
\newblock In \emph{2019 {Conference on Cognitive Computational Neuroscience}},
  2019.
\newblock \url{doi.org/10.32470/CCN.2019.1241-0}.

\bibitem[Rayner and Well(1996)]{RaynerWell1996}
K.~Rayner and A.~D. Well.
\newblock Effects of contextual constraint on eye movements in reading: {A}
  further examination.
\newblock \emph{Psychonomic Bulletin \& Review}, 3\penalty0 (4):\penalty0
  504--509, 1996.

\bibitem[Sukumaran et~al.(2022)Sukumaran, Houghton, and
  Kazanina]{SukumaranEtAl2022}
P.~Sukumaran, C.~Houghton, and N.~Kazanina.
\newblock Do {LSTMs} see gender? {Probing} the ability of {LSTMs} to learn
  abstract syntactic rules.
\newblock \emph{arXiv:2211.00153}, 2022.

\bibitem[Turing(1950)]{Turing1950}
A.~M. Turing.
\newblock Computing machinery and intelligence.
\newblock \emph{Mind}, 49:\penalty0 433–460, 1950.

\end{thebibliography}
\end{document}